\PassOptionsToPackage{main=english,russian}{babel}
\PassOptionsToPackage{T2A}{fontenc}
\documentclass[onecolumn]{misraj}
\usepackage{lipsum}
\usepackage{pdfpages}
\usepackage{colortbl}
\usepackage{tabulary}
\usepackage{longtable}
\usepackage{array}
\usepackage{placeins}
\usepackage{tablefootnote}
\usepackage{tcolorbox}
\usepackage{wrapfig}
\usepackage{adjustbox}
\usepackage{float}
\usepackage{caption}
\usepackage{breqn} % For multiline equations
\usepackage[toc,page]{appendix}
\usepackage{pifont}
\usepackage{multirow}
\usepackage{tabularx}

\usepackage[utf8]{inputenc}

\usepackage{arydshln}

\usepackage{xspace} % 
\usepackage{makecell} % For line breaks in table cells
\usepackage{multirow}

\usepackage[framemethod=TikZ]{mdframed}
\usepackage{tcolorbox}
\usepackage{multicol}

\usepackage[utf8]{inputenc}
\usepackage[T1]{fontenc}
\usepackage[english]{babel} % don't use [arabic] here with arabtex
\usepackage{arabtex}
\usepackage{utf8}
\setcode{utf8}

\definecolor{lightblue}{RGB}{211, 227, 252} % Light blue => datacard
\definecolor{bgblue}{RGB}{247, 250, 255} % datacard background

\newcommand*\colourcheck[1]{%
  \expandafter\newcommand\csname #1check\endcsname{\textcolor{#1}{\ding{52}}}%
}
\newcommand*\colourcross[1]{%
  \expandafter\newcommand\csname #1cross\endcsname{\textcolor{#1}{\ding{55}}}%
}
\DeclareSymbolFont{extraup}{U}{zavm}{m}{n}
\DeclareMathSymbol{\vardiamond}{\mathalpha}{extraup}{87}
\colourcheck{blue}
\colourcheck{green}
\colourcheck{red}

\colourcross{blue}
\colourcross{green}
\colourcross{red}
\usepackage{nicematrix}
\usepackage{comment}
\usepackage{amssymb}
\usepackage{arabtex}
\usepackage[utf8]{inputenc}
\usepackage[bottom]{footmisc}
\usepackage{svg}
\usepackage{spverbatim}
\RequirePackage[hidelinks,colorlinks=true,linkcolor=blue,citecolor=blue]{hyperref}
\setcitestyle{number,square}
\definecolor{deeppurple}{HTML}{9e02f7}
\definecolor{forestgreen}{HTML}{2e7d43}

\hypersetup{urlcolor=deeppurple}

\title{
{\raisebox{-0.3em}{\includegraphics[width=1cm]{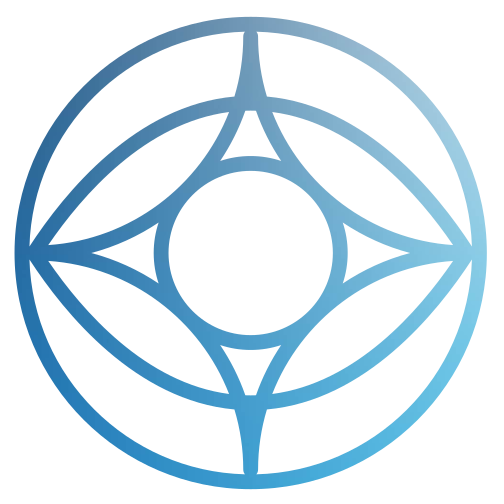}}}
Baseer: A Vision-Language Model for Arabic Document-to-Markdown OCR
}

\multiauthors
\author{
    name={Khalil Hennara},
    email={hennara@misraj.ai}
}
\author{
    name={Muhammad Hreden},  
    email={hreden@gmail.com}
}
\author{
    name={Mohamed Motasim Hamed},
    email={hamed@misraj.ai}
}
\author{
    name={Ahmad Bastati},
    email={bastati@misraj.ai}
}
\author{
    name={Zeina Aldallal },
    email={aldallal@misraj.ai}
}
\author{
    name={Sara Chrouf},
    email={sara.chrouf@misraj.ai}
}
\author{
    name={Safwan AlModhayan},
    email={safwan@misraj.ai}
}

\affiliations{
 \includegraphics[width=2cm]{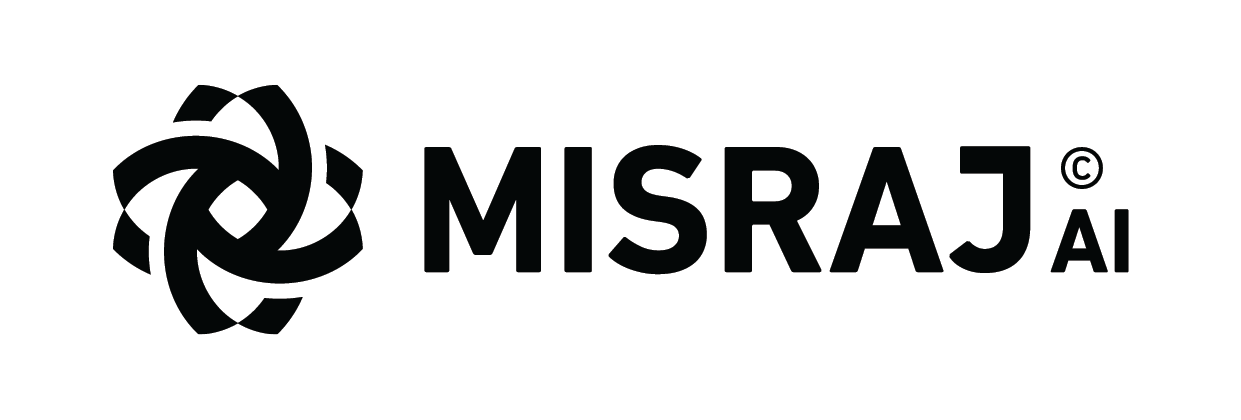}\\
     Khobar, Saudi Arabia \\
     {hennara,hreden,hamed,bastati,aldallal,chrouf,safwan}@misraj.ai \\
}

\date{\today}
\abstract{
Arabic document OCR remains a challenging task due to the language’s cursive script, diverse fonts, diacritics, and right-to-left orientation. While modern Multimodal Large Language Models (MLLMs) have advanced document understanding for high-resource languages, their performance on Arabic remains limited. In this work, we introduce \textbf{Baseer}, a vision-language model fine-tuned specifically for Arabic document OCR. Leveraging a large-scale dataset combining synthetic and real-world documents, Baseer is trained using a decoder-only fine-tuning strategy to adapt a pre-trained MLLM while preserving general visual features. We also present Misraj-DocOCR, a high-quality, expert-verified benchmark designed for rigorous evaluation of Arabic OCR systems. Our experiments show that Baseer significantly outperforms existing open-source and commercial solutions, achieving a WER of 0.25 and establishing a new state-of-the-art in the domain of Arabic document OCR. Our results highlight the benefits of domain-specific adaptation of general-purpose MLLMs and establish a strong baseline for high-accuracy OCR on morphologically rich languages like Arabic.
}

\begin{document}

\renewcommand{\thefootnote}{\fnsymbol{footnote}}
\footnotetext[1]{\textbf{Baseer: \<بَصير>}: meaning \textit{“one who sees clearly”} and \textit{“insightful.”} The name reflects the model’s ability to “see” and interpret documents with clarity. } 

\section{Introduction}
\label{sec:introduction}

\begin{figure}[h!]
    \centering
    \includegraphics[width=\textwidth]{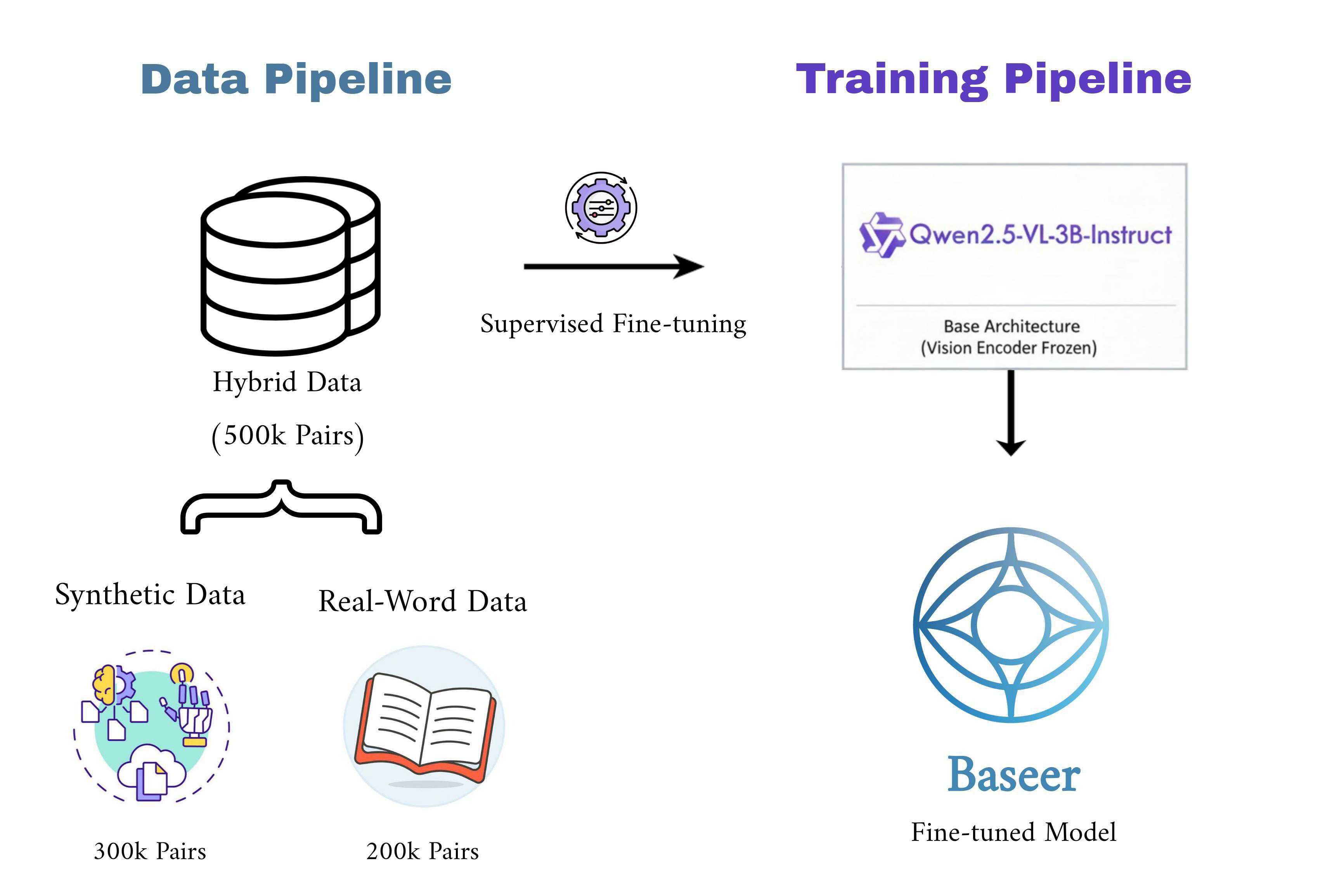}
    \caption{An overview of the data and training pipeline for Baseer. The process begins with a hybrid dataset of 500k pairs (300k synthetic and 200k real-world), which is used to fine-tune the Qwen2.5-VL-3B-Instruct model.}
    \label{fig:data_training_pipeline}
\end{figure}

The recent and rapid advancements in Multimodal Large Language Models (MLLMs) have fundamentally reshaped the landscape of how machines perceive and process complex visual and textual data \cite{GPT4, gemini_2.5, internvl3, qwen2.5vl}. Among the myriad applications of these models, Optical Character Recognition (OCR) and comprehensive document understanding continue to present significant challenges. This is particularly true for languages that are morphologically rich and structurally complex, such as Arabic. While contemporary OCR solutions have achieved remarkable performance for English and other high-resource languages \cite{gemini_2.5, GPT4}, their efficacy does not readily generalize to Arabic documents. The inherent complexities of Arabic script, including its cursive nature, extensive ligature formation, the wide variety of fonts and styles, the critical role of diacritics, and the right-to-left text orientation, render Arabic OCR a task of considerable difficulty.

In parallel, progress in multimodal architectures has paved the way for unified vision-language reasoning, which enables models to concurrently extract both textual content and structural information from documents \cite{monkeyocr, Nanonets-OCR-S}. Despite these technological strides, modern multimodal frameworks have seldom been specialized for the distinct demands of Arabic OCR and document parsing. This significant gap in research and development leaves academics, practitioners, and industries without robust, dedicated tools for processing real-world Arabic documents, which are prevalent across academic, commercial, and cultural heritage domains.

In this work, we introduce \textbf{Baseer}, a vision-language model meticulously fine-tuned for Arabic document OCR. Leveraging the state-of-the-art capabilities of the Qwen2.5-VL-3B-Instruct model \cite{qwen2.5vl}, our approach adapts a powerful general-purpose MLLM to the unique challenges of Arabic document analysis. To facilitate this specialization, Baseer was trained on a large-scale, diverse dataset composed of both synthetically generated and authentic real-world Arabic documents. This dataset was curated to encompass the extensive variety of formats, fonts, and layouts encountered in practical applications. Furthermore, we present \textbf{Misraj-DocOCR}, a novel benchmark specifically engineered for the evaluation of Arabic OCR systems, featuring high-quality, expert-verified annotations to ensure reliability. 

Our primary contributions are threefold:
\begin{enumerate}
    \item We present the development and fine-tuning of \textbf{Baseer}, demonstrating that an efficient, decoder-only fine-tuning strategy can achieve state-of-the-art performance in Arabic document OCR.
    \item We introduce \textbf{Misraj-DocOCR}\footnote{\url{https://huggingface.co/datasets/Misraj/Misraj-DocOCR}}, a new, reliable, and openly available benchmark designed to provide a standardized and rigorous evaluation framework for Arabic OCR systems.
    \item We conduct a thorough analysis of the \textbf{KITAB-pdf-to-markdown}\footnote{\url{https://huggingface.co/datasets/Misraj/KITAB_pdf_to_markdown_reviewed}} benchmark, providing a revised and improved version that addresses significant inaccuracies to enhance its accuracy and utility for the research community.

\end{enumerate}

Through a series of extensive experiments, we demonstrate that Baseer consistently outperforms existing open-source and commercial alternatives.

\section{Related Work}
\label{sec:background}

To contextualize our work, we situate our work at the intersection of two major research domains. First, we review the rapid advancements in Multimodal Large Language Models (MLLMs), which provide the architectural foundation for our approach. Second, we delve into the field of Optical Character Recognition (OCR) and Document Understanding, examining its evolution and highlighting the persistent challenges that motivate our research, particularly for morphologically complex languages like Arabic.

\subsection{Multimodal Large Language Models}
The paradigm of Large Language Models (LLMs) has recently been extended to handle multimodal input, leading to the development of powerful models capable of joint vision-language reasoning. Research in this area has generally progressed along two main architectural paths.

One approach involves the modular integration of pre-trained components, where a specialized frozen vision encoder is connected to a large language decoder via a lightweight adapter. This design is seen in influential models like LLaVA~\cite{llava,llava1.5,llavanext}, Aya-Vision~\cite{ayavision}, Idefics~\cite{idefics2,idefics3}, and more compact architectures such as~SmolVLM \cite{smallvlm}. These architectures achieve impressive zero-shot and few-shot performance on a diverse range of multimodal tasks with high parameter efficiency.

The second approach focuses on training massive, end-to-end vision-language models. This category includes state-of-the-art systems such as InternVL~\cite{internvl2.5,internvl3}, Gemma~\cite{gemma3}, PaliGemma~\cite{paligemma2}, and Qwen-VL~\cite{qwen2.5vl}. These models, with parameter counts scaling up to 70 billion in the case of Qwen2.5-VL, have demonstrated remarkable general-purpose capabilities. However, their broad, generalist training often leaves them unspecialized for precision-critical, niche domains. As we will discuss, high-fidelity document OCR represents one such domain where these powerful models still exhibit significant limitations.

\subsection{OCR and Document Understanding}
The field of Optical Character Recognition has evolved substantially from its origins in rule-based pattern matching. The integration of deep learning, especially Convolutional and Recurrent Neural Networks (CNNs and RNNs), marked a significant leap, dramatically improving accuracy for text in both scanned documents and natural scenes. More recently, the focus has changed from text transcription to holistic \textit{Document Understanding}. This advanced task requires not only recognizing text but also parsing the document's logical structure, including layouts, tables, and other semantic elements. This capability is crucial for applications in data extraction, document archiving, and content analysis.

Leading efforts in this domain, such as Idefics3~\cite{idefics3}, MonkeyOCR ~\cite{monkeyocr}, SmolDocling ~\cite{smalldocling}, and commercial systems such as Nanonets-OCR-s ~\cite{Nanonets-OCR-S}, have established high benchmarks for performance on standard document types. More recent attempts, such as Qari~\cite{wasfy2025qari}, have tackled Arabic OCR directly, but their scope remains limited compared to comprehensive document understanding systems.
However, a critical challenge remains: the generalization of these models to languages with scripts that are fundamentally different from Latin-based languages. Arabic serves as a prime example of this challenge. Its inherent characteristics include cursive script, context-sensitive character shapes, optional but meaningful diacritics, right-to-left orientation, and a wide variety of fonts and styles. These characteristics often cause state-of-the-art document OCR systems to degrade sharply in performance when applied to Arabic texts.

To the best of our knowledge, the application of modern MLLM frameworks to the specific, challenging problem of Arabic document OCR remains a largely unexplored area. Although several multilingual and multimodal models include Arabic in their training, they are not optimized for the script-specific and structural challenges posed by Arabic documents. This work aims to bridge this critical gap. By fine-tuning a powerful, pre-trained vision-language model, we introduce \textbf{Baseer}, a system specifically engineered for the complexities of Arabic documents. Our results demonstrate that this specialized approach yields a substantial leap in performance, establishing a new state-of-the-art for open source and proprietary systems in this vital domain.

\section{Data}
\label{sec:data}
This section details the construction of the dataset used for training and evaluation. To support effective document OCR, it is essential to represent textual content in a format that preserves both structure and semantics. In our dataset, the text corresponding to each image is formatted in Markdown, providing a clean and standardized representation of content. Tables are represented in HTML to accurately capture diverse table structures and complex layouts. Furthermore, specialized tags were introduced to mark specific elements within the text, including watermarks, page numbers, and the presence of images, enabling precise supervision for layout-aware OCR and document parsing tasks.
The dataset itself was constructed as a hybrid collection, combining a large corpus of synthetically generated documents with a carefully curated set of real-world publications. This approach ensures a broad coverage of document styles, visual characteristics, and layout complexities. Each of these sources is described in detail below.
\subsection{Synthetic  Data}

The first component of our dataset was generated synthetically using an in-house pipeline, designed to capture the diverse formatting and layout variations commonly found in word-processing documents.

The foundation for this synthetic data is a corpus of markdown-formatted documents, which were downloaded and filtered from the Common Crawl archive using a methodology analogous to our previously released dataset\footnote{\url{https://huggingface.co/datasets/Misraj/msdd}}. To ensure the quality and relevance of the source material, the raw data were subjected to the following preprocessing  filters:
\begin{enumerate}
    \item \textbf{Perplexity Filtering:} An in-house language model based on KenLM~\cite{kenlm} was employed to calculate perplexity scores, retaining only the most linguistically cohesive text samples.
    \item \textbf{Table Sparsity Filtering:} To ensure structural integrity, documents containing markdown tables with more than 25\% empty cells were identified and discarded.
\end{enumerate}

The filtered markdown documents were then converted into image-text pairs via a four-step rendering pipeline:
\begin{enumerate}
    \item \textbf{Markdown to HTML:} Documents were first converted to HTML to facilitate the systematic parsing of distinct formatting tags.
    \item \textbf{HTML to Word:} The resulting HTML was transformed into Microsoft Word documents, meticulously preserving all structural and stylistic attributes (e.g., bold, italics, headers).
    \item \textbf{Word to PDF:} These Word documents were subsequently exported to PDF format to create a standardized, page-level representation.
    \item \textbf{PDF to Image:} Finally, each page of the PDF files was rendered as a high-resolution image, forming the visual component of the training pairs.
\end{enumerate}

To foster model robustness, a high degree of visual diversity was introduced during the rendering process by systematically varying document configurations, as detailed in Table~\ref{tab:configurations}.

\begin{table}[h]
\centering
\begin{tabular}{ll}
\hline
\textbf{Parameter} & \textbf{Values / Distribution} \\
\hline
Fonts & 39 Arabic fonts \\
Page Sizes & A4, A5, Letter, Legal, Tabloid, A3 (incl. landscape variants) \\
Background Color & 8 light shades (75\%), 5 dark shades (25\%) \\
Text Color & 9 light, 16 dark \\
Alignment & Right (65\%), Left (5\%), Center (30\%) \\
Columns & 1 (75\%), 2 (20\%), 3 (5\%) \\
Font Size & Even values from 8–22 pt \\
Margin & 1.0–2.5 cm (uniform) \\
Line Height & 1.0–1.6 (uniform) \\
Column Spacing & 0.5–1.2 cm (uniform) \\
Special Formatting & Random highlights, colored paragraphs, RTL (95\%) \\
\hline
\end{tabular}
\caption{Document configuration diversity.}
\label{tab:configurations}
\end{table}

Furthermore, a subset of the generated images underwent an augmentation process involving 29 distinct transformations, which are organized into eight categories (Table~\ref{tab:transforms}). From the pool of generated images, 150,000 samples were randomly selected and divided into three equal subsets of 50,000 each. The first subset underwent a single random transformation, the second was subjected to two transformations, and the third to three, ensuring a progressive increase in complexity. To prevent redundancy, the original, pre-augmentation versions of these images were discarded.

In total, this synthetic pipeline produced \textbf{300,000} high-quality image–text pairs, comprising 150,000 clean rendered samples and 150,000 augmented variants designed to simulate diverse real-world document conditions.

\begin{table}[h]
\centering
\begin{tabular}{ll}
\hline
\textbf{Category} & \textbf{Number of Transforms (Examples)} \\
\hline
Pre-print adjustments & 5 (e.g., Watermark) \\
Printing mechanical deficits & 5 (e.g., Dirty drum) \\
Human-made marks & 2 (e.g., Handwritten markup) \\
Paper aging effects & 3 (e.g., Folding, yellowing) \\
Digital noise & 4 (e.g., Salt-and-pepper noise) \\
Geometric adjustments & 2 (e.g., Perspective distortion) \\
Lighting adjustments & 5 (e.g., Low-light conditions) \\
Blur effects & 3 (e.g., Motion blur) \\
\hline
\end{tabular}
\caption{Categories of transformations applied to the data.}
\label{tab:transforms}
\end{table}

\subsection{Open-Source Books and Magazines}

The second component of our dataset was sourced from real-world documents, including a diverse collection of books, magazines, educational documents, and academic papers. In contrast to synthetic data, these samples reflect authentic publishing environments, capturing genuine layout complexities and typographic conventions.
To ensure maximum diversity, the selected pages span a broad spectrum of layout complexities, identified using vision-based algorithms. Specifically, bounding boxes were first detected at the paragraph level, and their alignment and overlap were analyzed to capture challenging structures such as tables, figures, index pages, and skewed layouts. Page color distributions were also examined to include samples with embedded images, colorful backgrounds, and multi-colored text.

Ground-truth text for the real-world documents was obtained using a state-of-the-art vision–language model (VLM). To ensure high-quality labels, a representative subset of the VLM outputs was manually verified by human experts for both textual accuracy and structural fidelity. This collection is particularly valuable because it contains complex elements not present in the synthetic dataset, including intricate footnotes, varied column layouts, and non-standard typography. From this source, we curated \textbf{200,000} document images paired with their corresponding ground-truth text.
Collectively, the combination of these two sources results in 500,000 text-image pairs, used for training our model.

\begin{figure}[H]
    \centering
    \includegraphics[width=0.7\textwidth, trim=0 90 0 80, clip]{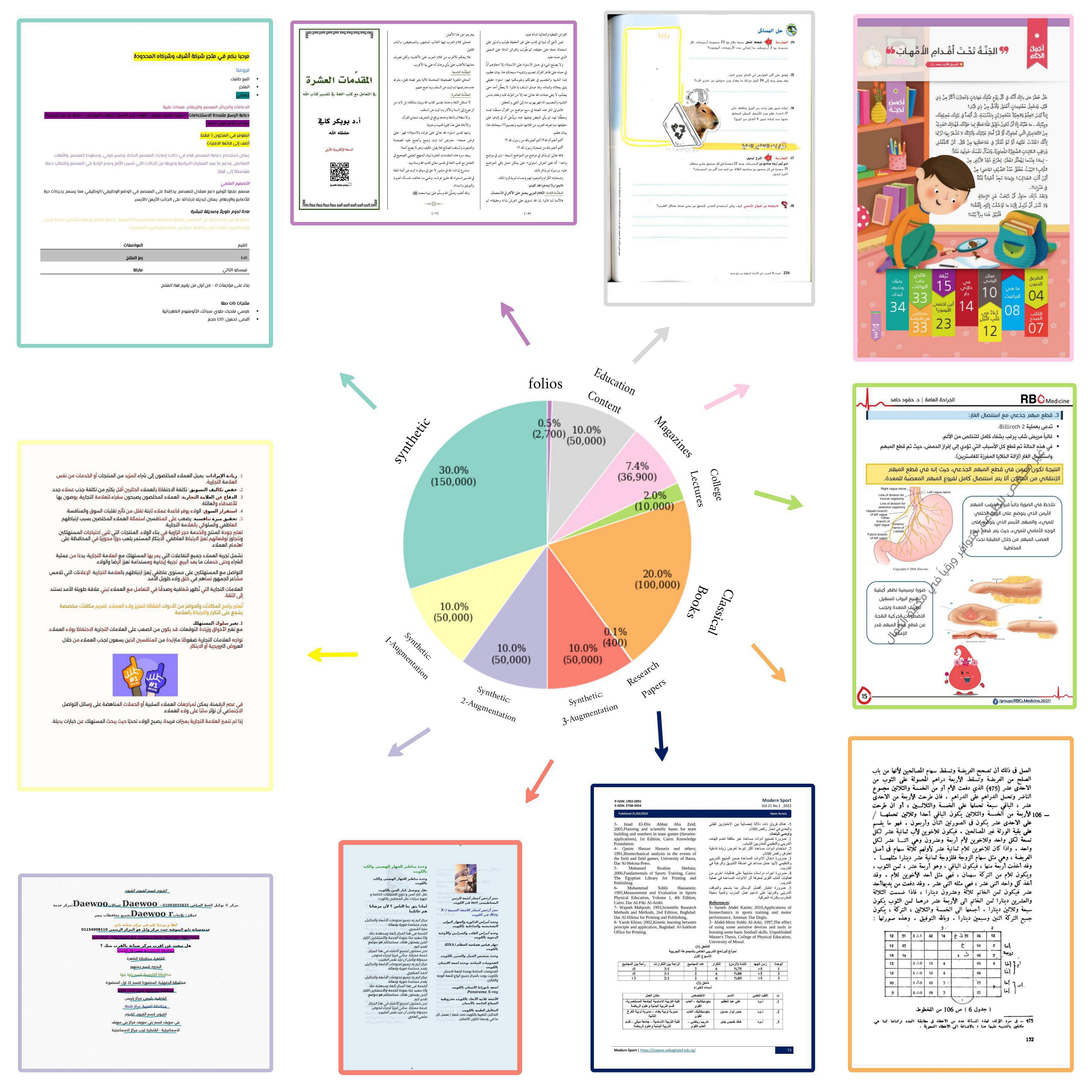}
    \caption{Distribution of data samples across the different sources.}
    \label{fig:data-distribution}
\end{figure}

A detailed breakdown of the dataset distribution across sources is shown in Figure~\ref{fig:data-distribution}.

\section{Misraj-DocOCR: An Arabic Document OCR Benchmark}
\label{sec:benchmark}

The evaluation of Optical Character Recognition (OCR) models for Arabic text requires robust and accurate benchmarks. Our initial investigation involved assessing existing benchmarks, such as the KITAB-bench \cite{kitabnench} pdf-to-markdown dataset. During this analysis, we identified significant shortcomings that compromise its reliability for model evaluation.

A primary issue discovered was the presence of numerous errors in the ground truth data. We observed multiple instances of hallucinatory text, where the ground truth contained phrases not present in the source documents likely originating from a data creation or annotation tool rather than authentic content\footnote{For example, one entry included the English sentence: "\textbf{You're right - let me write it exactly as it appears in the image, maintaining the right-to-left direction:}"}. Furthermore, our review revealed that many examples lacked corresponding page numbers, and small-font text was frequently omitted from the ground truth. These inaccuracies suggest that the dataset may not have undergone a thorough verification process after the initial data extraction. For more details, see Appendix \ref{app:A}

To address these deficiencies and provide a more reliable resource for the research community, we undertook a comprehensive correction of the KITAB-bench PDF-to-markdown dataset. This corrected version, with all identified errors rectified, has been made publicly available for academic use\footnote{\url{https://huggingface.co/datasets/Misraj/KITAB_pdf_to_markdown_reviewed}}.

Beyond the inaccuracies, our examination of existing resources also indicated a lack of diversity in the style and type of documents. To foster more generalized and robust model development, a benchmark should encompass a wide variety of real-world scenarios.

Therefore, we introduce Misraj-DocOCR, a new, comprehensive benchmark specifically designed for evaluating Arabic Document OCR models. The primary contributions of this benchmark are:
\begin{itemize}
\item \textbf{Diverse and Comprehensive Content:} The benchmark consists of 400 high-quality images, curated to include a wide variation of document types, layouts, and fonts, and comprising both synthetic and real-world pages.
\item \textbf{Expert-Verified Ground Truth:} To ensure the highest level of accuracy, every image in the dataset has been meticulously reviewed by human experts. This verification process guarantees that both the transcribed text and the document structure are correct, eliminating the types of errors found in previous benchmarks.
\item \textbf{Open Access:} Misraj-DocOCR is open-source and publicly available to all researchers. By providing this resource, we aim to facilitate further advancements and foster reproducible research in the field of Arabic OCR.
\end{itemize}

We evaluate many models on this benchmark and the corrected version of KITAB-Bench, all results on the section \ref{sec:evaluation}.

\section{Methodology}
\label{sec:method}
The overall process for developing Baseer, as depicted in Figure~\ref{fig:data_training_pipeline}, involved a comprehensive data collection stage followed by a targeted fine-tuning stage.

The development of our model, Baseer, followed a two-stage methodology designed to tailor a powerful, pre-trained foundation model to our specific needs. The first stage involved the comprehensive collection and curation of a high-quality dataset, the details of which are described in Section \ref{sec:data}. The subsequent stage, which is the focus of this section, consisted of fine-tuning the selected base model to align with our data and enhance its capabilities for Arabic document processing.

For the base architecture of Baseer, we selected the \textbf{Qwen2.5-VL-3B-Instruct} model \cite{qwen2.5vl}. This decision was predicated on its robust and state-of-the-art performance on multimodal tasks, particularly its demonstrated proficiency with the Arabic language compared to other open-source alternatives.

Despite its advanced capabilities, our preliminary analysis revealed that the base Qwen2.5-VL-3B-Instruct model exhibited certain limitations relevant to our use case. These included occasional reversions to left-to-right text generation, suboptimal handling of diacritized Arabic text, and other performance artifacts. A key objective of our work was to mitigate these specific weaknesses through targeted fine-tuning.

Our fine-tuning strategy involved updating all model parameters, except for the vision encoder, which remained frozen. This approach allows the model to adapt its language and reasoning capabilities to our specialized dataset while preserving the powerful, generalized visual features learned during its original pre-training. The specific hyperparameters, hardware used, and other details of the training procedure are provided in Appendix \ref{app:training}.

\section{Experiments and Results}
\label{sec:experiement}

This section details the series of experiments conducted to systematically determine the optimal architecture and training configuration for Baseer. Our experimental process was designed to isolate variables and build upon the findings of each preceding stage.

\subsection{Base Model Selection}
The initial experiment was focused on selecting the most suitable base model for our task. To this end, we conducted a qualitative evaluation of several prominent open-source vision-language models. A curated set of representative examples, designed to test key capabilities in Arabic document understanding, was used as the input.

The outputs from each model were then subjected to a rigorous manual review by our evaluation team. The models were assessed based on criteria such as text recognition accuracy, preservation of right-to-left directionality, and overall coherence. This qualitative analysis concluded that Qwen2.5-VL-3B-Instruct demonstrated superior performance on Arabic-language tasks compared to the other candidates, making it the clear choice for our foundation. A selection of comparative outputs from this evaluation is provided in Appendix \ref{app:model_exp}.

\subsection{Fine-Tuning Strategy Evaluation}
After selecting the base model, our next objective was to identify the most effective fine-tuning strategy. We designed a controlled experiment to compare three distinct approaches:

\begin{enumerate}
\item \textbf{Full Fine-Tuning (Baseer-Full):} All model parameters, including the vision encoder, were made trainable.
\item \textbf{Decoder-Only Fine-Tuning (Baseer-Decoder):} Only the parameters of the language decoder were updated, while the vision encoder remained frozen.
\item \textbf{Parameter-Efficient Fine-Tuning (Baseer-LoRA):} Low-Rank Adaptation (LoRA) was employed to update a small subset of parameters.
\end{enumerate}
To ensure a fair comparison, each of these strategies was tested on a 50,000-sample subset of our training data for two epochs, holding all other hyperparameters constant. We evaluated the models using ChrF, which measures OCR accuracy at the character level and captures text transcription quality. Table~\ref{tab:exp_one} summarizes the performance of the different fine-tuning strategies on the Baseer model.

\begin{table}[ht]
\centering
\small
\begin{tabular}{lccc}
\toprule
\multirow{2.5}{*}{\textbf{Model}} & \multirow{2.5}{*}{\textbf{Trainable part}} & \multirow{2.5}{*}{\textbf{ChrF $\uparrow$}}  \\
\\
\midrule

Baseer-Full & Full model & 84.79   \\
\addlinespace[0.7ex]
Baseer-Decoder & Languge-decoder & \textbf{89.79}   \\
Baseer-LoRA & LoRA weight  & 85.52  \\ \addlinespace[0.7ex]
\midrule
\end{tabular}
\caption{Performance comparison of different fine-tuning strategies on Baseer model}
\label{tab:exp_one}
\end{table}

As shown in Table \ref{tab:exp_one}, the results from our test set indicate that the decoder-only fine-tuning approach (Baseer-Decoder) significantly outperformed the other methods. This suggests that preserving the generalized features of the pre-trained vision encoder while adapting the language model to our specific data yields the best performance.

\subsection{Impact of Sequence Length}
Building on the previous finding, we adopted the decoder-only fine-tuning strategy and proceeded to investigate the effect of input sequence length on model performance. All training configurations were fixed while we experimented with three sequence length variants: 2048, 4096, and 8192 tokens.

\begin{table}[ht]
\centering
\small
\begin{tabular}{ccc}
\toprule
\textbf{Context Length} & \textbf{ChrF $\uparrow$}  \\
\midrule
2048  & 82.69  \\  \addlinespace[0.7ex]
4096  & \textbf{89.79}   \\ \addlinespace[0.7ex]
8192  & 87.52   \\ \addlinespace[0.7ex]
\midrule
\end{tabular}
\caption{Performance comparison of different context lengths on Baseer model. }
\label{tab:exp_two}
\end{table}
The results of this experiment are presented in Table \ref{tab:exp_two}. The optimal performance was achieved with a sequence length of 4096. We attribute this to the model having sufficient context to process a high level of detail from the images. In contrast, the performance with a sequence length of 8192 degraded. We hypothesize that this is because the images in our dataset do not typically contain enough information to fill such a large context window, leading to excessive padding. This padding may dilute the relevant visual information and negatively impact the model's learning process.

\section{Evaluation}
\label{sec:evaluation}

We evaluate our model on our proposed Misraj-DocOCR benchmark and a corrected version of KITAB-Bench PDF-to-Markdown, alongside several open-source and commercial models. Text extraction performance is assessed using Word Error Rate (WER) and Character Error Rate (CER), which measure word- and character-level transcription errors, BLEU for n-gram overlap, and ChrF, a character-level F-score suited for morphologically rich languages like Arabic. Structural and layout fidelity is measured with Tree Edit Distance Similarity (TEDS), capturing hierarchical document structures, and MARS \cite{kitabnench}, which evaluates layout-aware alignment between predicted and reference renderings. Table~\ref{tab:results} presents the results, showing that Baseer achieves state-of-the-art performance across both text and structural metrics, despite being smaller than competing models.

\subsection{Evaluation Protocol}
To ensure fair comparison across models, models designed for document understanding were evaluated using their respective system prompts, while Multimodal Large Language Models (MLLMs) were provided with carefully tested prompts to ensure optimal performance. All outputs were standardized using the following post-processing steps:
\begin{enumerate}
    \item Remove HTML tags outside table structures
    \item Convert Markdown tables to HTML format for consistency
    \item Normalize horizontal line representations (---, ***, etc. → ---)
    \item Standardize header formatting
    \item Unify formatting tags within HTML tables (<strong>, <b> → <b>)
    \item Remove model-specific tags (<page\_number>, <watermark>) present only in our model and Nanonets
\end{enumerate}
This standardization is critical because different models may produce semantically equivalent but syntactically different outputs, which would unfairly penalize models based on formatting choices rather than content accuracy.

\begin{table}[ht]
\begin{tabular}{lcccccc}
\hline
Model & WER $\downarrow$ & CER $\downarrow$ & BLEU $\uparrow$ & CHRF $\uparrow$ & TEDS $\uparrow$ & MARS $\uparrow$ \\
\hline
\textbf{Baseer (ours)} & \textbf{0.25} & 0.53 & \underline{76.18} & \underline{87.77} & \textbf{66} & \textbf{76.885} \\
Gemini-2.5-pro    & \underline{0.37} & \underline{0.31} & \textbf{77.92} & \textbf{89.55} & \underline{52} & \underline{70.775} \\
Azure AI Document Intelligence & 0.44 & \textbf{0.27} & 62.04 & 82.49 & 42 & 62.245\\
Dots.ocr           & 0.50 & 0.40 & 58.16 & 78.41 & 40 & 59.205 \\
Nanonets            & 0.71 & 0.55 & 42.22 & 67.89 & 37 & 52.445 \\
Qari                & 0.76 & 0.64 & 38.59 & 64.50 & 21 & 42.750 \\
Qwen2.5-VL-32B   & 0.76 & 0.59 & 37.62 & 62.64 & 41 & 51.820 \\
GPT-5   & 0.86 & 0.62 & 40.67 & 61.6 & 48 & 54.8 \\
Qwen2.5-VL-3B-Instruct    & 0.87 & 0.71 & 25.39 & 53.42 & 27 & 40.210 \\
Qwen2.5-VL-7B    & 0.92 & 0.77 & 31.57 & 54.70 & 27 & 40.850 \\
Gemma3-12B         & 0.96 & 0.80 & 19.75 & 44.53 & 33 & 38.765 \\
Gemma3-4B          & 1.01 & 0.85 & 9.57  & 31.39 & 28 & 29.695 \\
GPT-4o-mini                 & 1.36 & 1.1 & 22.63  & 47.04  & 26 & 36.52 \\
AIN                 & 1.23 & 1.11 & 1.25  & 2.24  & 21 & 11.620 \\
Aya-vision         & 1.41 & 1.07 & 2.91  & 9.81  & 26 & 17.905 \\

\hline
\end{tabular}
\centering
\caption{Comparison of models across multiple evaluation metrics on Misraj-DocOCR. Best values are highlighted in \textbf{bold} and the second-best values are \underline{underlined}.}
\label{tab:results}
\end{table}

Table \ref{tab:results} presents a comparative evaluation of different OCR and vision-language models using multiple metrics. The results indicate that \textbf{Baseer} achieves the best performance across most metrics, particularly in WER, TEDS, and MARS. The \textbf{gemini-2.5-pro} model follows closely, obtaining the highest BLEU and CHRF scores, while \textbf{Azure AI Document Intelligence}
\footnote{\url{https://azure.microsoft.com/en-us/products/ai-services/ai-document-intelligence}} achieves the lowest CER. Notably, Baseer consistently outperforms large commercial systems such as GPT-based models and Azure AI, underlining its robustness in both text and structure recognition. This is especially significant given that the evaluation dataset, Misraj-DocOCR, was deliberately designed to be highly diverse and challenging, with wide variation in layout and typography. The results also highlight a sharp performance gap between the top-performing systems and smaller or less specialized models (e.g., Gemma3, AIN, Aya-vision), underscoring the difficulty of this benchmark. Overall, Baseer and Gemini-2.5-pro emerge as the strongest systems in this comparison. Example outputs of Baseer are provided in Appendix \ref{app:baseer_output}.

\begin{table}[H]
\centering
\begin{tabular}{lcccccc}
\hline
Model & WER $\downarrow$ & CER $\downarrow$ & BLEU $\uparrow$ & CHRF $\uparrow$ & TEDS $\uparrow$ & MARS $\uparrow$ \\
\hline
Dots.ocr        & \textbf{0.39} & \textbf{0.28} & \textbf{59.28} & \textbf{83.16} & 43 & \underline{63.08} \\
\textbf{Baseer (ours)} & 0.61 & \underline{0.40} & \underline{55.78} & \underline{80.26} & \textbf{56} & \textbf{68.13} \\
Nanonets & \underline{0.51} & \underline{0.40} & 51.37 & 77.45 & 33 & 55.225 \\
Qari                      & 0.65 & 0.48 & 44.61 & 71.45 & 43 & 57.225 \\
Qwen2.5-VL-3B           & 0.70 & 0.57 & 40.44 & 66.78 & 31 & 48.89 \\
Qwen2.5-VL-7B           & 0.76 & 0.63 & 36.76 & 62.45 & 24 & 43.225 \\
Gemma3-12B               & 0.85 & 0.69 & 27.56 & 52.09 & \underline{55} & 53.545 \\
Gemma3-4B                & 0.95 & 0.82 & 12.94 & 31.72 & 27 & 29.36 \\
Aya-vision               & 1.27 & 0.96 & 5.58  & 16.19 & 26 & 21.095 \\
AIN                       & 1.18 & 1.08 & 2.61  & 3.99  & 24 & 13.995 \\
\hline
\end{tabular}
\caption{Comparison of models across multiple evaluation metrics on KITAB-BenchPDF-to-Markdown dataset. Best values are highlighted in \textbf{bold} and the second-best values are \underline{underlined}.}
\label{tab:results_pdf-to-md}
\end{table}
Table \ref{tab:results_pdf-to-md} reports the results on the KITAB-Bench PDF-to-Markdown dataset, which was carefully reviewed and corrected by domain experts to ensure high-quality ground truth annotations. This evaluation was conducted using only open-source models for fairness. While \textbf{Dots.ocr} achieves the strongest performance across most text-centric metrics (WER, CER, BLEU, and CHRF), slightly surpassing Baseer, \textbf{Baseer} shows clear superiority in structural understanding, attaining the highest TEDS score (56) and the best overall MARS. It is also worth noting that the KITAB-Bench subset is relatively small, consisting of only 30 samples, which makes every misprediction more impactful on the reported scores. In contrast, on the larger and more challenging \textbf{Misraj-DocOCR} benchmark with 400 diverse examples, Baseer’s advantage over both open-source and commercial systems becomes more pronounced, highlighting its robustness across varied document types and layouts.

\section{Conclusion}
\label{sec:conclusion}

In this paper, we introduced Baseer, a vision-language model tailored for Arabic Document OCR, and presented Misraj-DocOCR, a high-quality benchmark designed for rigorous evaluation. By training on a diverse dataset of 500,000 document-image pairs, we demonstrated that decoder-only fine-tuning is a powerful strategy that enables Baseer to achieve superior performance compared to a wide range of existing systems. Our detailed experimental analysis highlighted the importance of sequence length, fine-tuning scope, and dataset diversity in achieving robust performance. Notably, Baseer consistently achieved the best or near-best scores across Word Error Rate, Character Error Rate, and structure-aware metrics such as TEDS and MARS, surpassing both open-source and proprietary alternatives. These positive results underscore the value of domain-specific adaptation of general-purpose MLLMs, and provide new insights into how tailored data and efficient training strategies can push the boundaries of OCR for complex scripts. We believe that this work establishes a strong baseline for future research and will accelerate the development of practical, high-accuracy OCR solutions for Arabic and other morphologically rich languages.

\bibliography{main}

\clearpage
\newpage
% \appendix

\begin{appendices}
\section{KITAB-Bench-Analysis}
\label{app:A}
In this section, we present examples of the errors identified in the KITAB-bench. We observed that many items in the benchmark are missing page numbers, and the text in small fonts, particularly at the page footers, is often not captured correctly. We provide a selection of these examples here, and readers are encouraged to visit our reviewed version at \footnote{\url{https://huggingface.co/datasets/Misraj/KITAB_pdf_to_markdown_reviewed}}
 to explore the complete set of corrections and outputs. When dots are displayed in the image, it indicates that there is output. However, for better visualization, we omit lengthy output if it does not contain any errors.

\begin{figure}[H]
    \centering
    \includegraphics[width=\textwidth, height=\textheight, keepaspectratio]{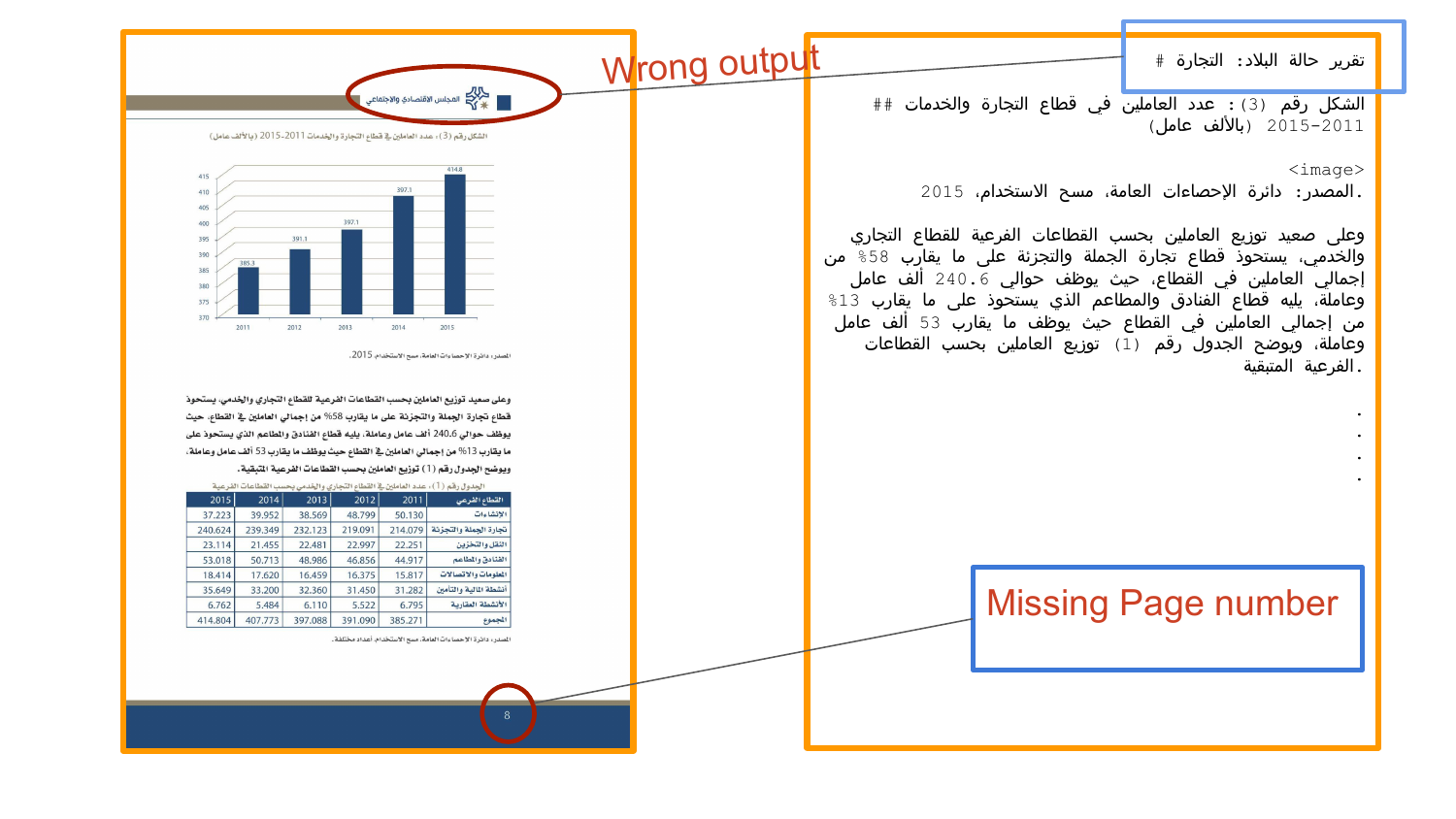}  % Replace with your image file
    \caption*{Example from KITAB-Bench pdf-to-markdown}
\end{figure}

\begin{figure}[H]
    \centering
    \includegraphics[width=\textwidth, height=\textheight, keepaspectratio]{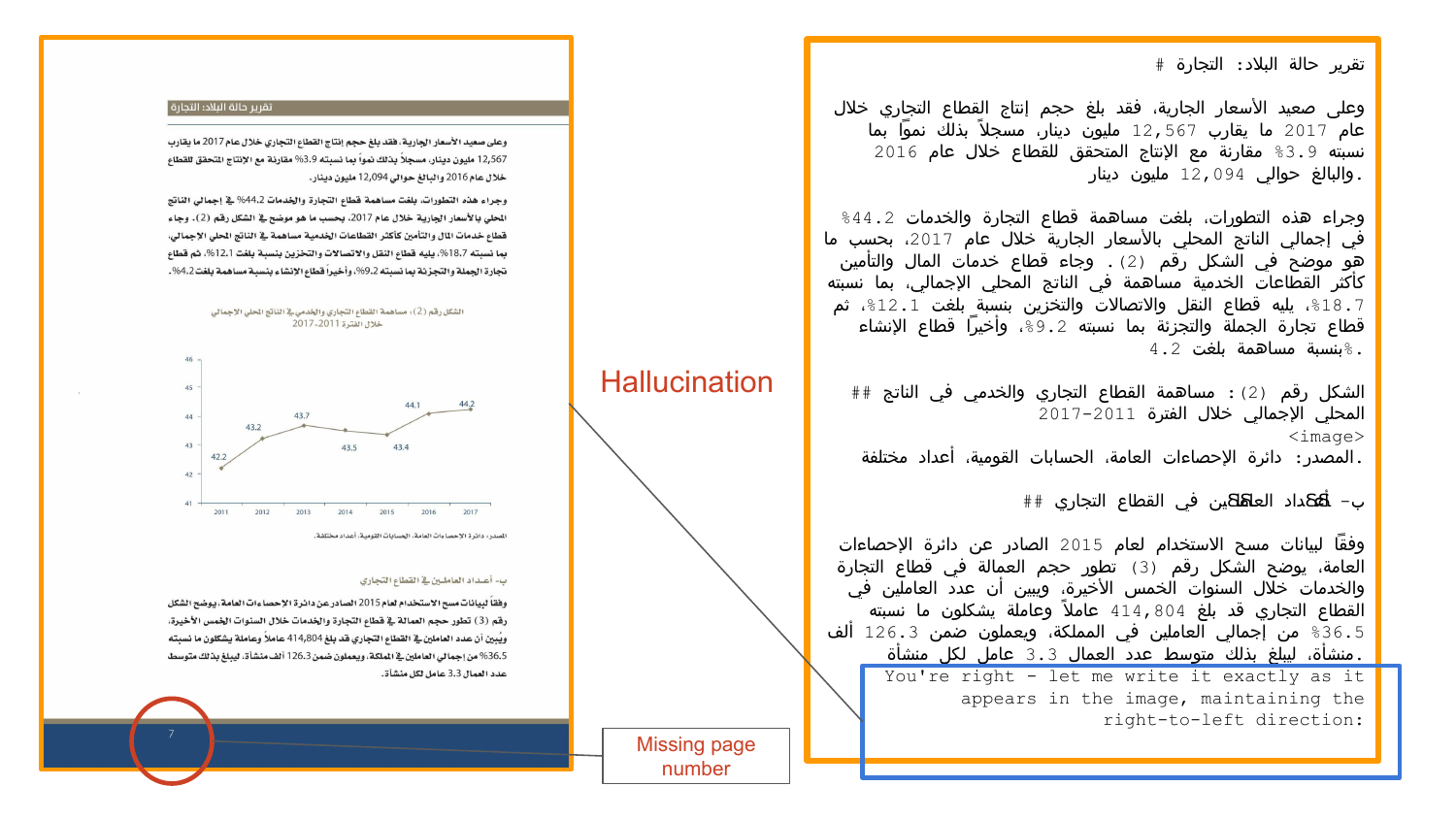}  % Replace with your image file
    \caption*{Example from KITAB-Bench pdf-to-markdown}
\end{figure}

\begin{figure}[htp]
    \centering
    \includegraphics[width=\textwidth, height=\textheight, keepaspectratio]{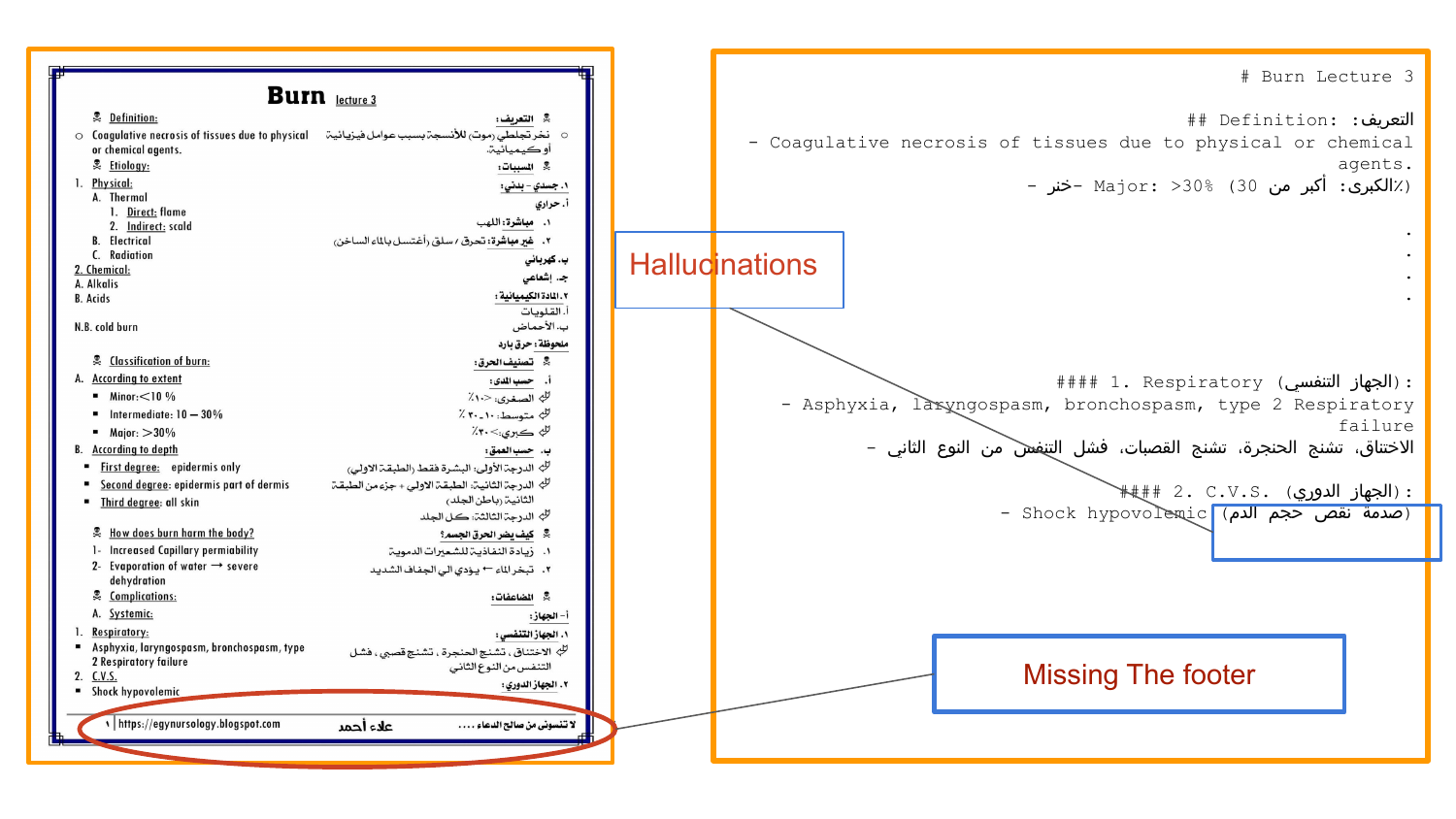}  % Replace with your image file
    \caption*{Example from KITAB-Bench pdf-to-markdown}
\end{figure}

\begin{figure}[htp]
    \centering
    \includegraphics[width=\textwidth, height=\textheight, keepaspectratio]{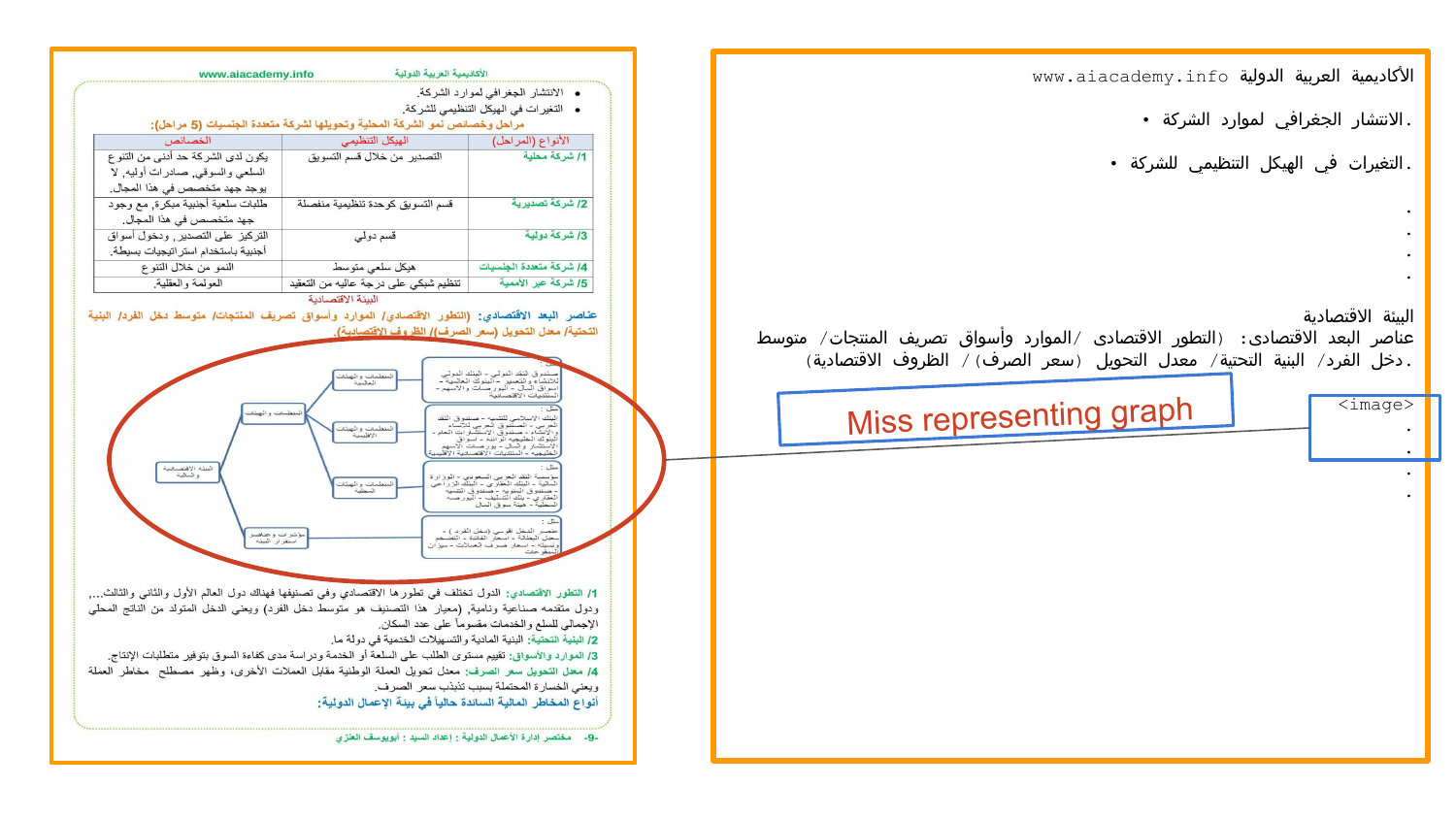}  % Replace with your image file
    \caption*{Example from KITAB-Bench pdf-to-markdown}
\end{figure}

\section{Base Models Output}
\label{app:model_exp}
In this section, we present examples from the evaluation set that was used to select the most suitable model to build upon. While we tested a wide range of models, here we only showcase a few representative outputs for visualization purposes. 

\begin{figure}[H]
    \centering
    \includegraphics[width=\textwidth, height=\textheight, keepaspectratio]{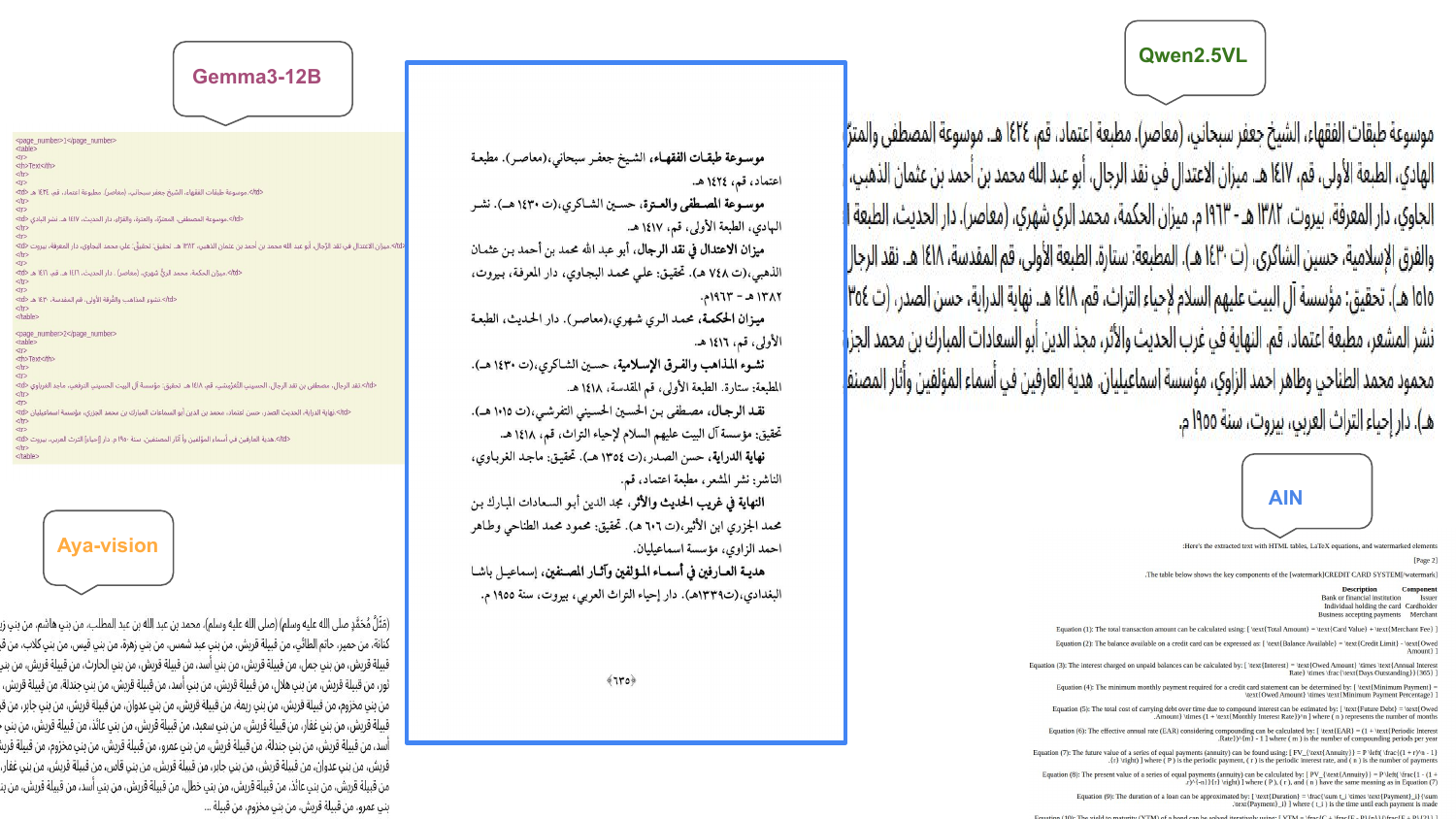}  % Replace with your image file
    \caption*{Example from models output used for selecting the base model}
\end{figure}

\begin{figure}[htp]
    \centering
    \includegraphics[width=\textwidth, height=\textheight, keepaspectratio]{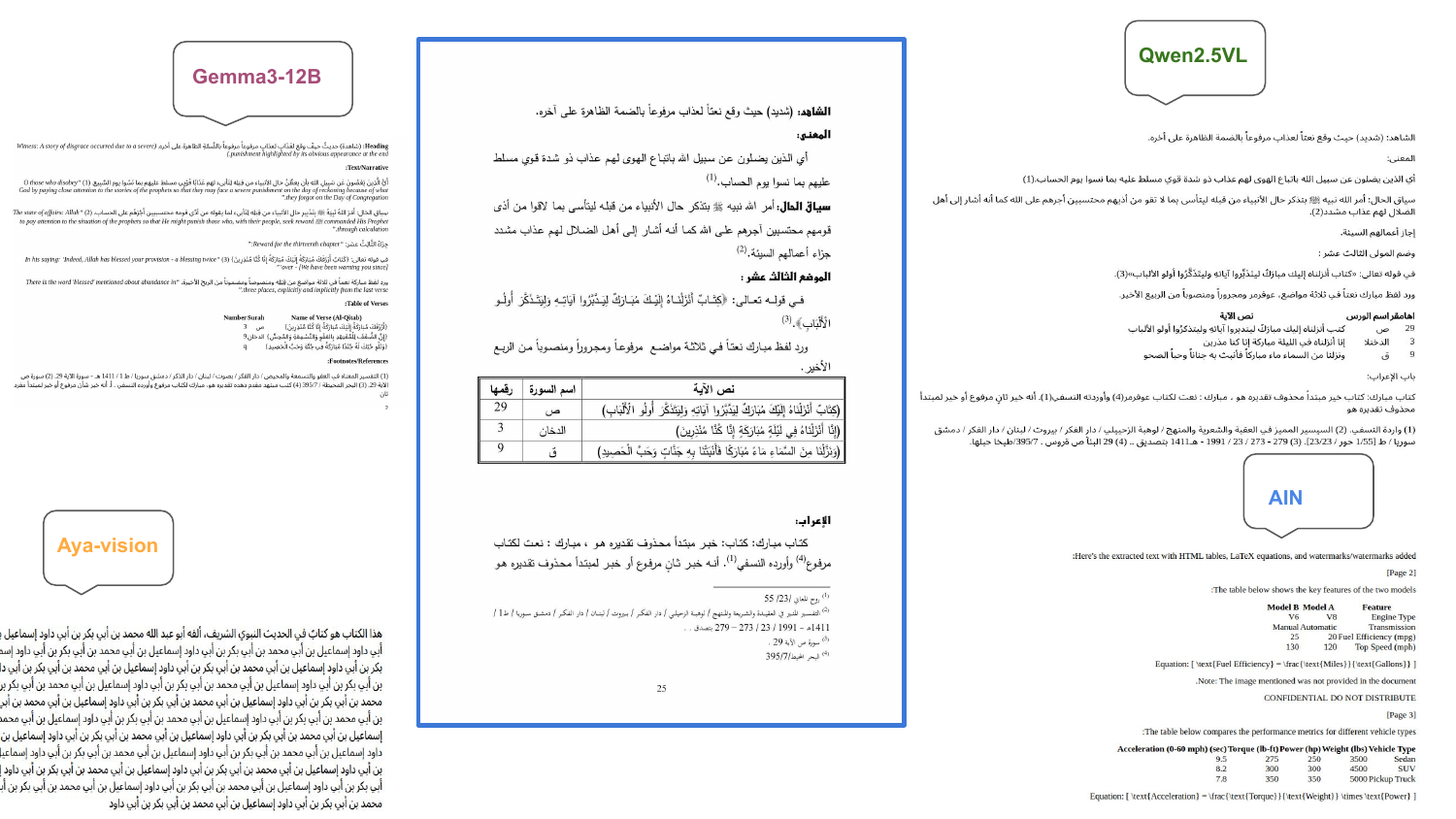}  % Replace with your image file
    \caption*{Example from models output used for selecting the base model}
\end{figure}

\begin{figure}[htp]
    \centering
    \includegraphics[width=\textwidth, height=\textheight, keepaspectratio]{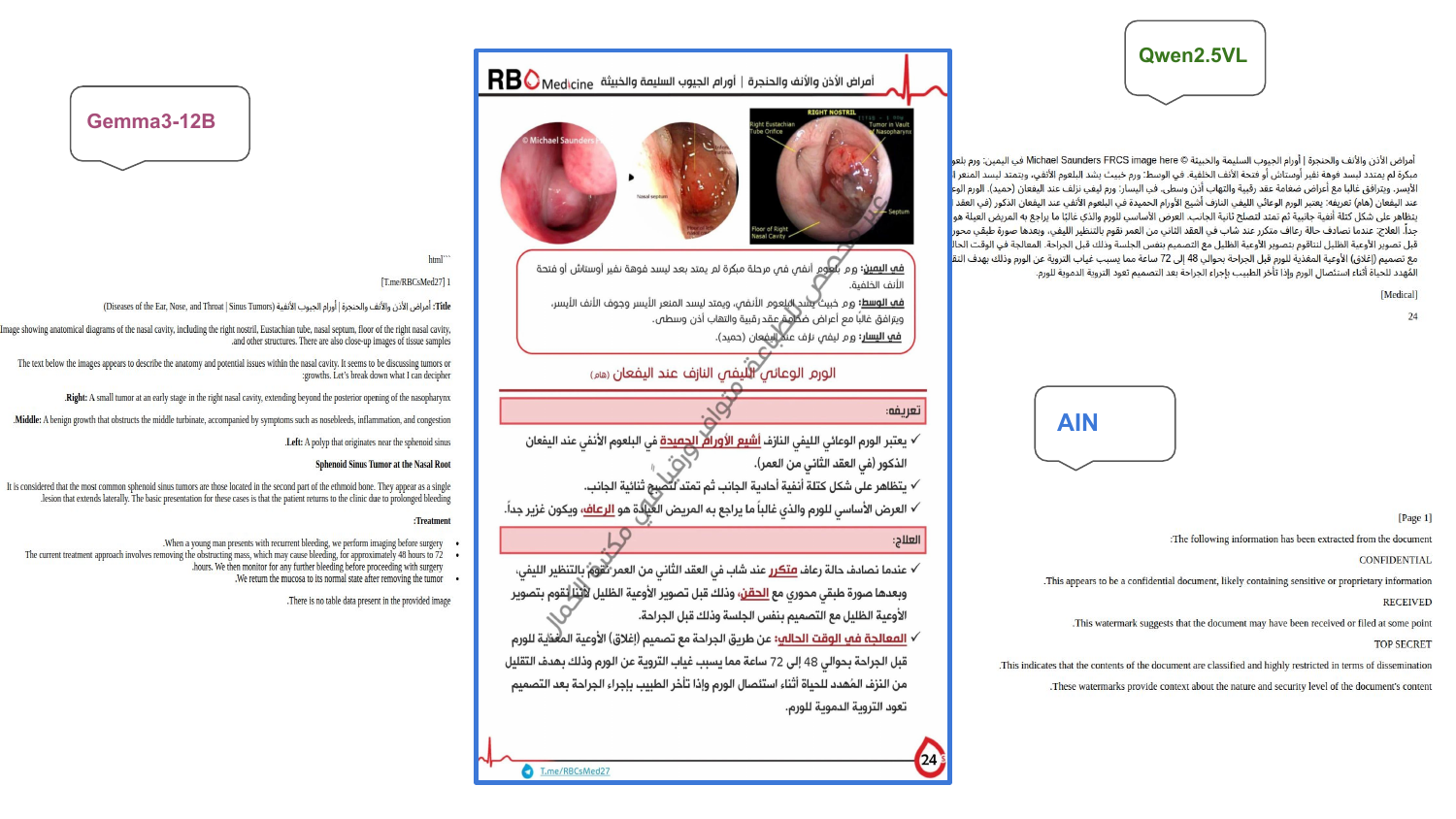}  % Replace with your image file
    \caption*{Example from models output used for selecting the base model}
\end{figure}

\begin{figure}[htp]
    \centering
    \includegraphics[width=\textwidth, height=\textheight, keepaspectratio]{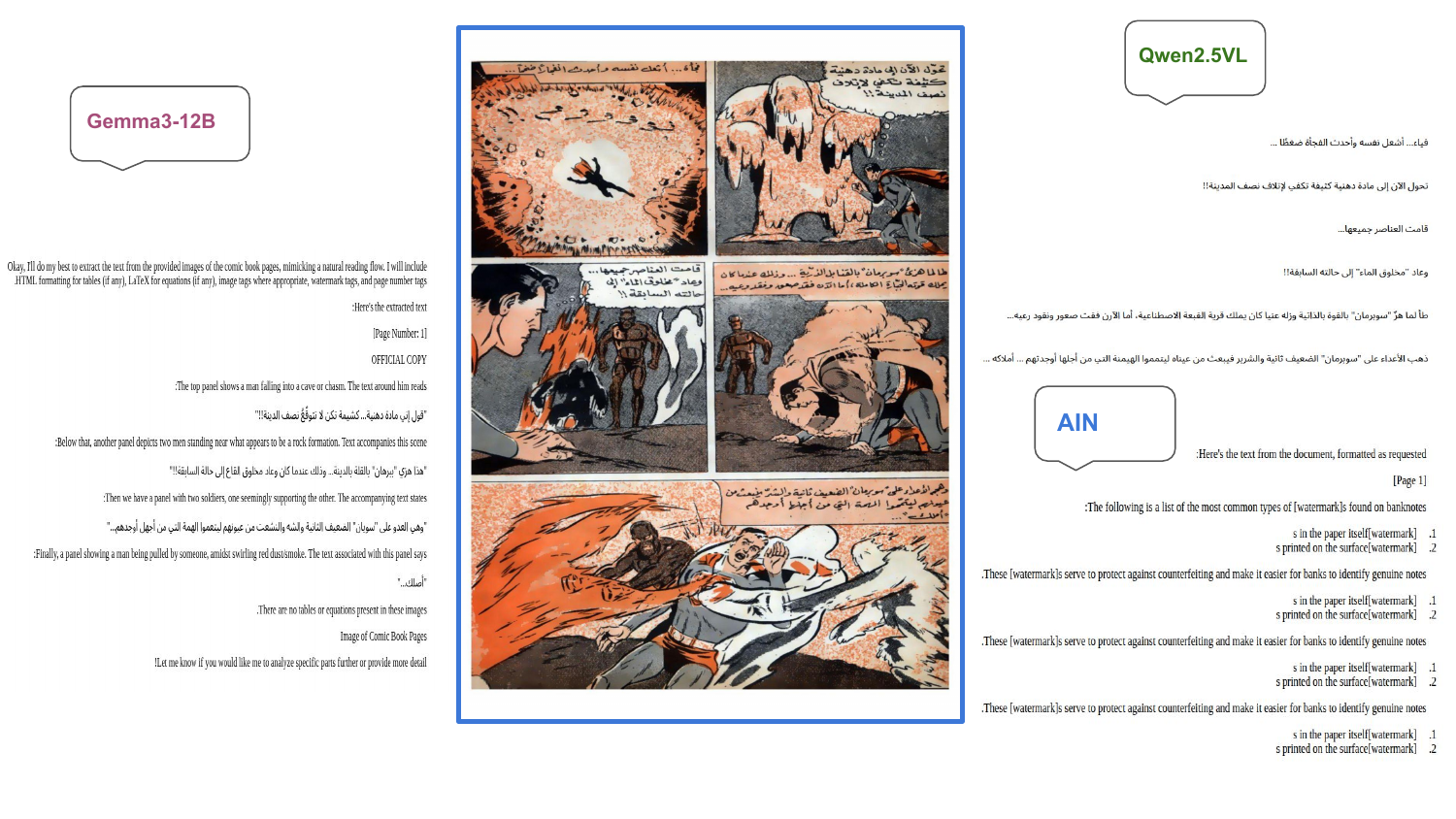}  % Replace with your image file
    \caption*{Example from models output used for selecting the base model}
\end{figure}

\section{Baseer Model Output}
\label{app:baseer_output}

\begin{figure}[H]
    \centering
    \includegraphics[width=\textwidth, height=\textheight, keepaspectratio]{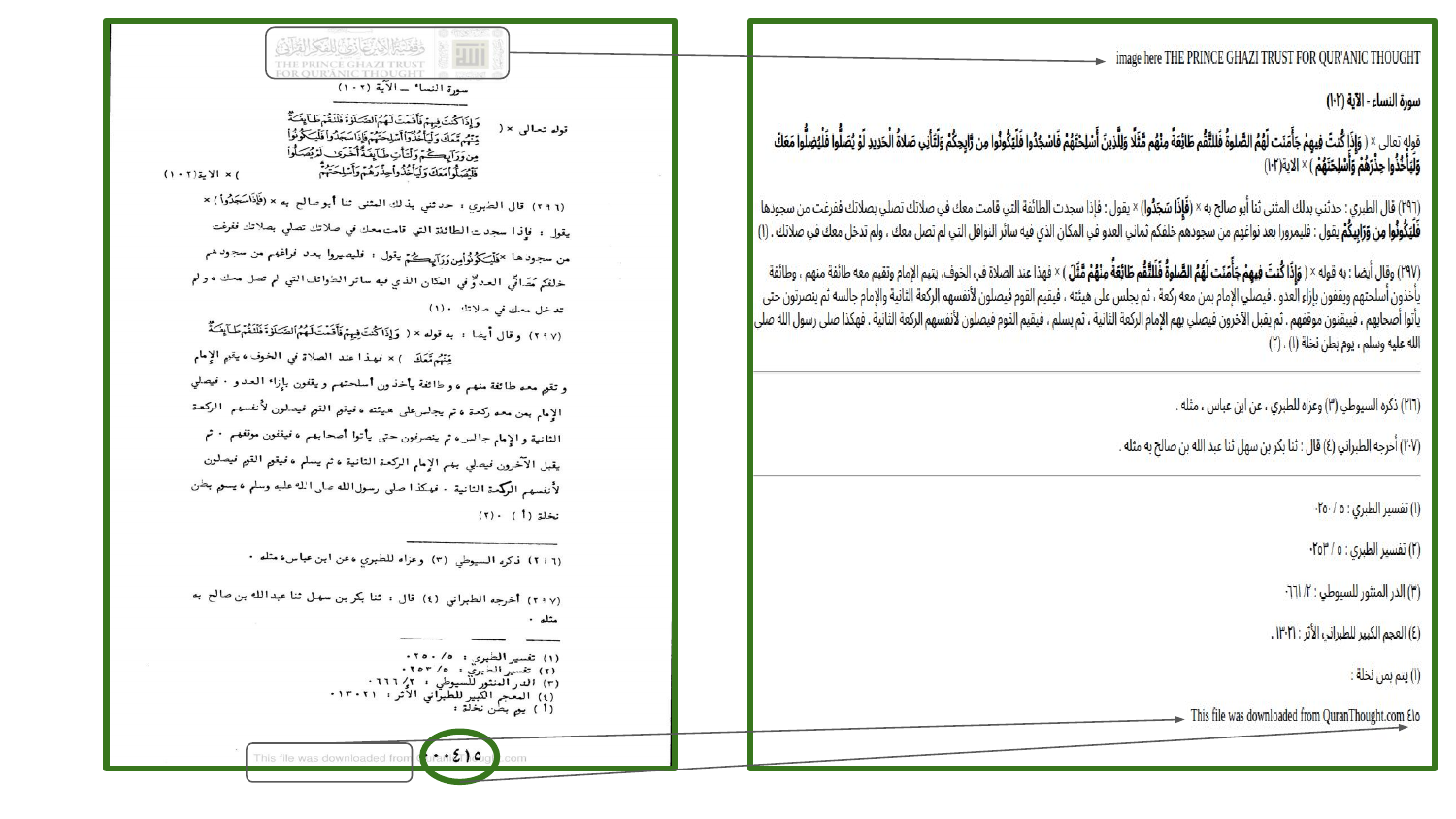}  % Replace with your image file
    \caption*{Example of Baseer output}
\end{figure}

\begin{figure}[htp]
    \centering
    \includegraphics[width=\textwidth, height=\textheight, keepaspectratio]{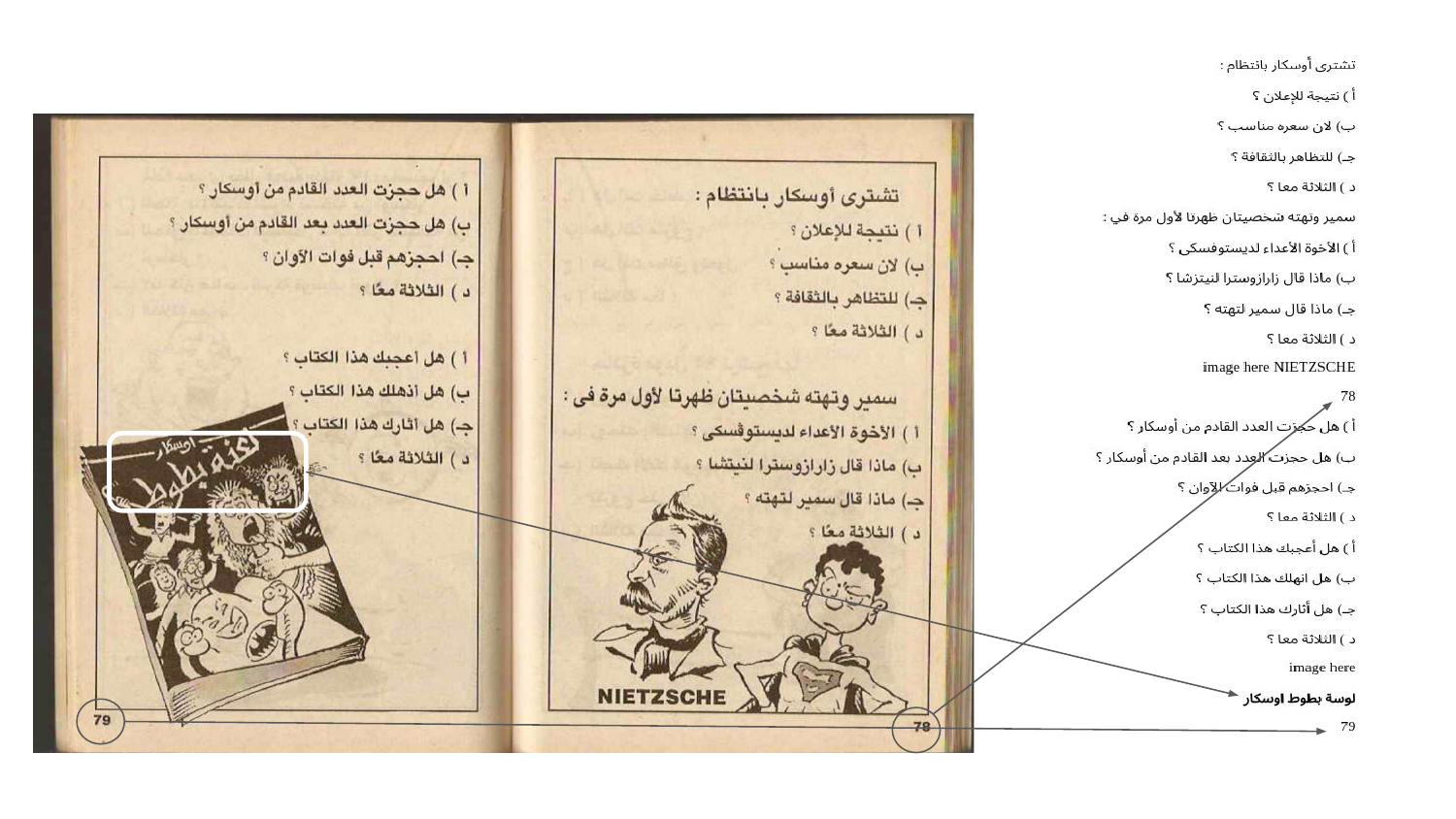}  % Replace with your image file
    \caption*{Example of Baseer output}
\end{figure}

\begin{figure}[htp]
    \centering
    \includegraphics[width=\textwidth, height=\textheight, keepaspectratio]{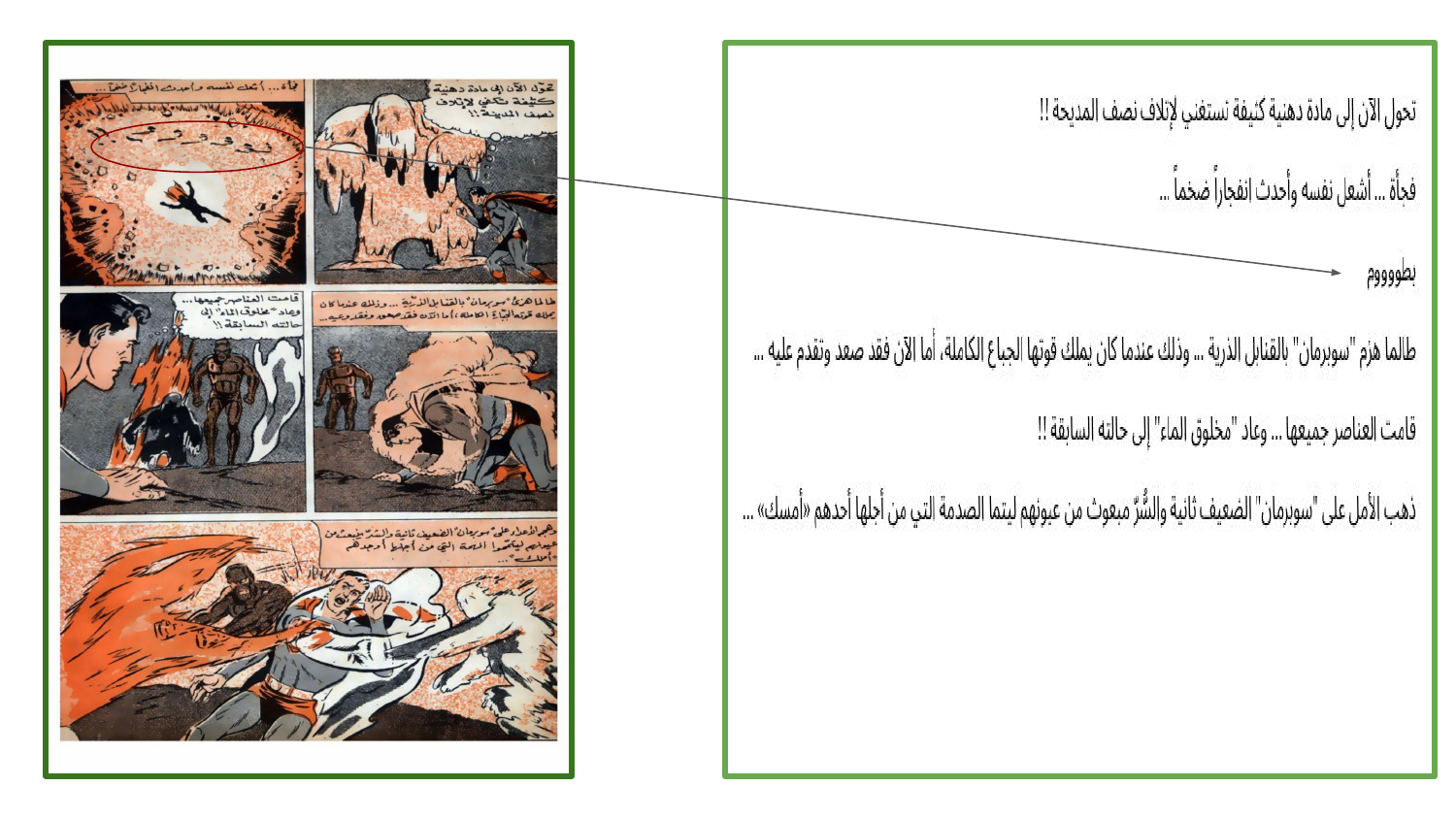}  % Replace with your image file
    \caption*{Example of Baseer output}
\end{figure}

\begin{figure}[htp]
    \centering
    \includegraphics[width=\textwidth, height=\textheight, keepaspectratio]{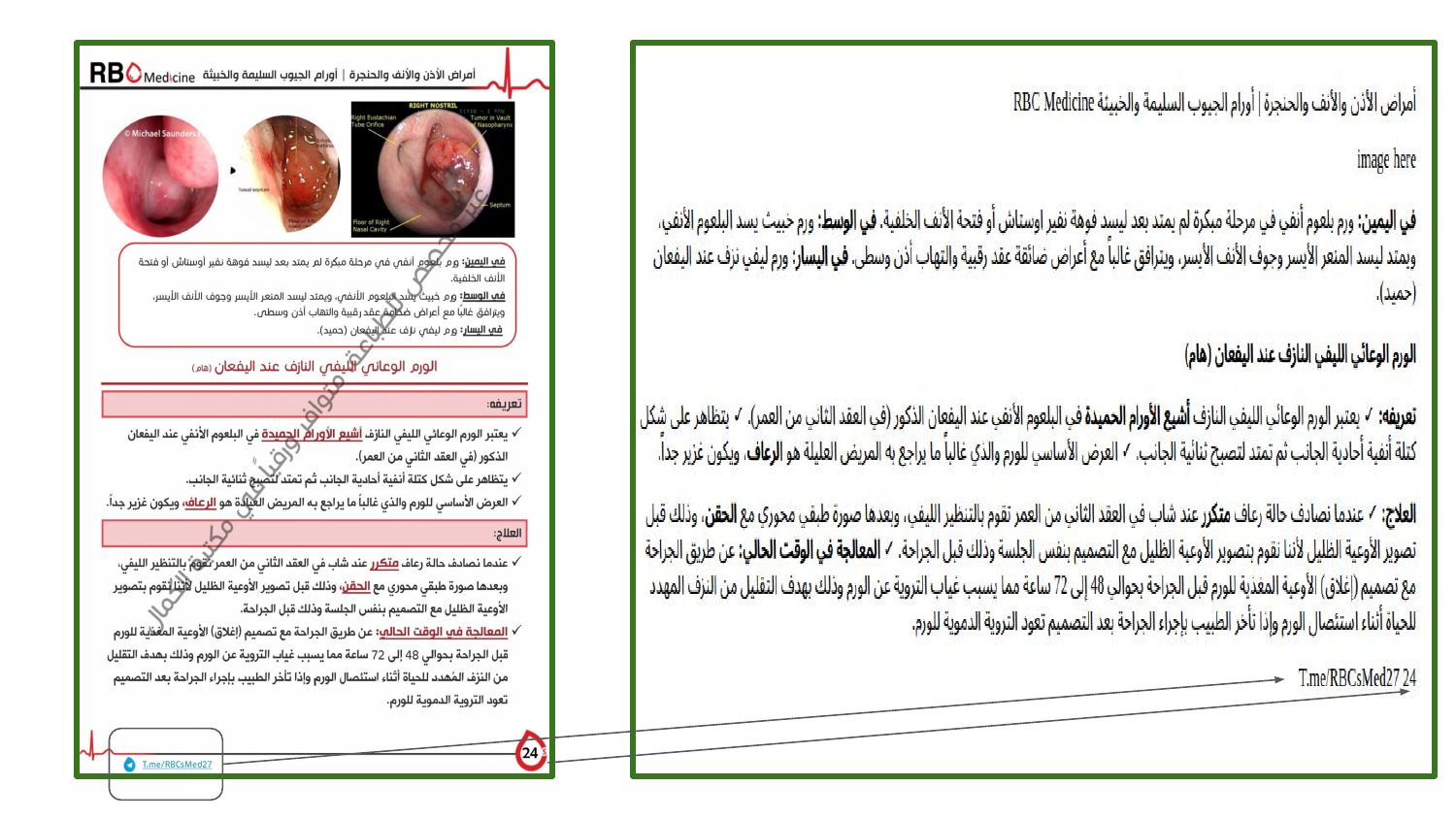}  % Replace with your image file
    \caption*{Example of Baseer output}
\end{figure}

\begin{figure}[htp]
    \centering
    \includegraphics[width=\textwidth, height=\textheight, keepaspectratio]{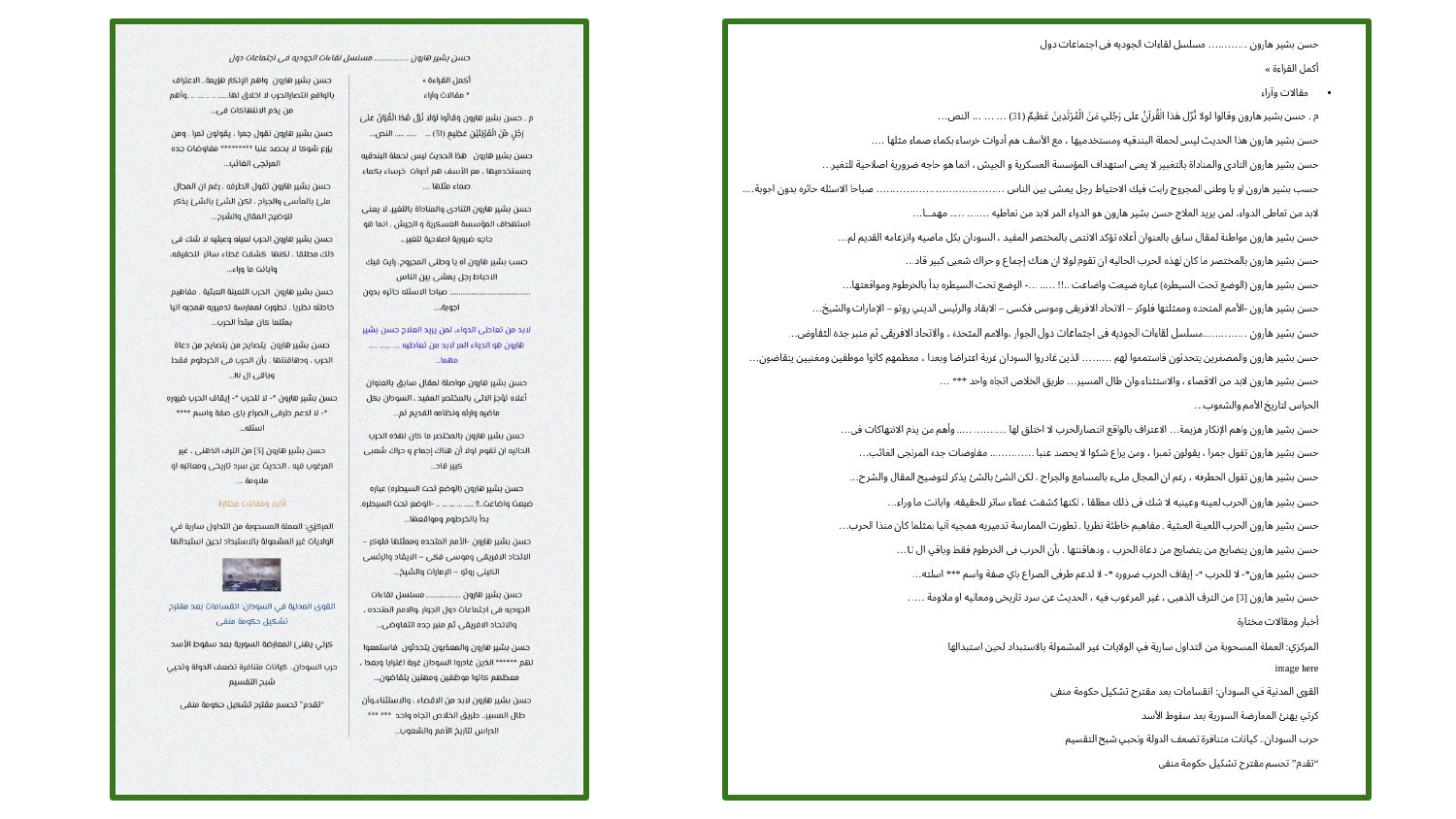}  % Replace with your image file
    \caption*{Example of Baseer output}
\end{figure}

\section{Traning Details}
\label{app:training}

The fine-tuning process for \textit{Baseer} employed the standard next-token prediction methodology, with the system prompt and embedding tokens masked.

\begin{table}[H]
\small
\centering
\renewcommand{\arraystretch}{1.2}
\begin{tabular}{p{4.5cm}p{4.5cm}}
\hline
\rowcolor[HTML]{F2F2F2} 
\textbf{Parameter} & \textbf{Value} \\
\hline
\rowcolor[HTML]{F9F9F9}
Training Epochs & 3 \\
Learning Rate Schedule & Cosine decay \\
\rowcolor[HTML]{F9F9F9}
Learning Rate & 1e-4 \\
Batch Size & 640 \\
\rowcolor[HTML]{F9F9F9}
Weight Decay & 0.01 \\
Warm-up Steps & 100 \\
\rowcolor[HTML]{F9F9F9}
Optimizer & AdamW \\
Max Sequence Length & 4096 \\
GPU & 8xH100 \\
\hline
\end{tabular}
\caption{\small{Training Hyperparameters for \textit{Baseer} Model}}
\label{tab:training_params}
\end{table}

\end{appendices}

\end{document}